\documentclass{article}

\usepackage{microtype}
\usepackage{graphicx}
\usepackage{subcaption}
\usepackage{booktabs} 

\usepackage{hyperref}
\usepackage{tcolorbox}

\usepackage[accepted]{icml2026}

\usepackage{amsmath}
\usepackage{amssymb}
\usepackage{mathtools}
\usepackage{amsthm}
\usepackage{enumitem}

\usepackage{multirow}
\usepackage[table]{xcolor}

\usepackage{graphicx}
\usepackage{adjustbox}
\usepackage{array}
\usepackage{graphicx}
\usepackage{caption}
\usepackage{xcolor}
\usepackage{pifont}
\definecolor{dino}{RGB}{227,240,249}
\definecolor{lightgray}{RGB}{240,240,240}

\usepackage[capitalize,noabbrev]{cleveref}

\theoremstyle{plain}

\theoremstyle{definition}

\theoremstyle{remark}

\usepackage[textsize=tiny]{todonotes}

\icmltitlerunning{MM-Snowball: Evaluating and Mitigating Hallucination Snowballing in Multimodal Multi-Turn Dialogue}

\begin{document}
\twocolumn[
  \icmltitle{MM-Snowball: Evaluating and Mitigating Hallucination\\ Snowballing in Multimodal Multi-Turn Dialogue}



  \icmlsetsymbol{equal}{*}

  \begin{icmlauthorlist}
    \icmlauthor{Yue Jiang}{equal,fdu}
    \icmlauthor{Xue Jiang}{equal,sust,hkbu}
    \icmlauthor{Lihua Zhang}{fdu}
    \icmlauthor{Zhiqiang Wang}{hkust}
    \icmlauthor{Yuhang Lu}{hw}
    \icmlauthor{Peng Wang}{hw} \\
    \icmlauthor{Bo Han}{hkbu}
    \icmlauthor{Feng Zheng}{sust}
    \icmlauthor{Dingkang Yang}{fdu,cuhk}
  \end{icmlauthorlist}

  \icmlaffiliation{fdu}{College of Intelligent Robotics and Advanced
Manufacturing, Fudan University}
  \icmlaffiliation{sust}{Southern University of Science and Technology}
  \icmlaffiliation{hkbu}{TMLR Group, Hong Kong Baptist University}
  \icmlaffiliation{cuhk}{MM Lab, CUHK}
  \icmlaffiliation{hkust}{HKUST}
  \icmlaffiliation{hw}{RAMS Lab, Huawei Technologies Co., Ltd.}

  \icmlcorrespondingauthor{Feng Zheng}{f.zheng@ieee.org}
  \icmlcorrespondingauthor{Dingkang Yang}{dickenyang@cuhk.edu.hk}

  \icmlkeywords{Multimodal, Dialogue, Hallucination}

  \vskip 0.3in
]



\printAffiliationsAndNotice{\icmlEqualContribution}

\begin{abstract}
Multimodal large language models (MLLMs) demonstrate remarkable visual understanding, yet their reliability in interactive settings is severely undermined by {hallucination snowballing}: a phenomenon where initial errors amplify across conversational turns, leading to a collapse in coherence. This failure reveals a fundamental vulnerability where models progressively neglect visual grounding in favor of over-relying on polluted textual history. Existing benchmarks are predominantly confined to single-turn VQA, which fail to capture the complex dynamics of error propagation in long-horizon interactions. To address this, we introduce {\textit{MM-Snowball}}, the first benchmark for fine-grained diagnosis of hallucination snowballing within dialogues. 
Extensive evaluation shows that our benchmark poses a significant challenge even to advanced MLLMs and reveals the inefficacy of existing mitigation methods designed for single-turn VQA. To counteract this degradation, we propose {\textit{Conflict-Aware Visual Rectification}} (CAVR). This training-free method mitigates snowballing through a synergistic dual-mechanism that refreshes visual grounding at the representation level and rectifies output distributions at the logit level, effectively re-anchoring the model to visual facts. Experiments demonstrate that CAVR achieves state-of-the-art performance, offering a promising path toward more reliable interactive AI. Data and code are available at: https://frenkie-chiang.github.io/MM-Snowball
\end{abstract}
\section{Introduction}
\label{sec:intro}

Multimodal large language models (MLLMs) have rapidly progressed on tasks such as visual question answering, captioning, and instruction-following. Fueled by large-scale pretraining and instruction tuning, these models can generate fluent, semantically rich descriptions and reason about images in natural language \cite{liu2023improvedllava,bai2025qwen25}. However, despite strong performance on static, single-turn benchmarks, MLLMs are known to suffer from multimodal hallucinations \cite{wang2025coin,leng2024mitigating,lin2026medcausalx,zhu2026medeyes}. In real-world applications, users typically interact with MLLMs through multi-turn dialogue, not isolated single-turn prompts. In this setting, hallucinations can be more pernicious \cite{cao2024visdiahalbench,zhong2024mmhalsnowball,luo2024halludial}. Once an incorrect object, attribute, or relation is introduced in an early turn, the model may repeatedly refer back to it, treat it as factual, and build increasingly complex inferences on top of it. Errors thus snowball over turns, leading to responses that are internally coherent but progressively detached from the visual evidence. We refer to this phenomenon as hallucination snowballing in multimodal dialogue.

Despite its significance, a systematic understanding of hallucination snowballing remains elusive due to two critical limitations in existing research \cite{zhong2024mmhalsnowball,sun2024mmhalbench,jiang2026danmakutppbench,li2023halueval}:
(1) most current benchmarks are confined to single-turn tasks or simplistic two-turn ``caption-then-question'' patterns. Such setups fail to capture the long-range error propagation and the non-linear dynamics of how a discrete perceptual slip evolves into a systemic cognitive delusion in realistic, long-horizon dialogues.
(2) Existing multi-turn datasets mostly use synthesized or edited images. They also lack a rigorous classification of how various noise sources, including misleading user premises and prior hallucinated responses, drive the snowballing effect. Consequently, we still lack a diagnostic framework to evaluate model resilience across progressive stages of interaction.

\begin{table*}[t]
\caption{\textbf{Comparison of hallucination benchmarks.} \textit{YN} and \textit{OE} denote yes-no and open-ended questions, respectively, and \textit{MCQ} stands for multiple-choice questions. \emph{Hallu-relative Questions} indicate whether the benchmark explicitly incorporates hallucination-inducing queries. \emph{Fake Image} indicates whether edited or synthesized images are employed instead of native visual inputs from real world to provoke artificial hallucinations. \emph{Hallu-relative Metrics} denote whether the benchmark defines dedicated evaluation metrics tailored to hallucination assessment. Enabled by our proposed \textit{AHTS} method, the dialogue in MM-Snowball unfolds progressively and can be readily extended to additional dialogue turns (beyond 6 turns).}

\renewcommand{\arraystretch}{1.2}
\resizebox{\textwidth}{!}{%

\begin{tabular}{lcccccccc}
\hline
\textbf{Benchmark/ Dataset}          & \textbf{Publication} & \textbf{Questions} & \textbf{\begin{tabular}[c]{@{}c@{}}Question \\ Type \end{tabular}} & \textbf{\begin{tabular}[c]{@{}c@{}}Hallu-relative\\ Question\end{tabular}} & \textbf{\begin{tabular}[c]{@{}c@{}}w/o\\ Fake Image\end{tabular}} & \textbf{\begin{tabular}[c]{@{}c@{}}Hallu-relative\\ Metrics\end{tabular}} & \textbf{\begin{tabular}[c]{@{}c@{}}Dialogue\\ Turns\end{tabular}} & \textbf{\begin{tabular}[c]{@{}c@{}}Progressive\\ Dialogue\end{tabular}} \\ \hline
\rowcolor{dino} \multicolumn{9}{c}{\textit{\textbf{Hallucination Benchmark}}}  \\ 
POPE~\cite{li2023evaluating}                                 & EMNLP'23           & 3,000               & YN                     & \textcolor{red}{\ding{55}}                                                                         & \textcolor{green}{\ding{51}}              & \textcolor{red}{\ding{55}}                                                               & \textcolor{red}{\ding{55}}                                                        & \textcolor{red}{\ding{55}}                                                                                 \\
MMVP~\cite{tong2024mmvp}                                 & CVPR'24            & 150                & YN/MCQ                     & \textcolor{green}{\ding{51}}                                                                         & \textcolor{red}{\ding{55}}              & \textcolor{red}{\ding{55}}                                                               & \textcolor{red}{\ding{55}}                                                        & \textcolor{red}{\ding{55}}                                                                                 \\
HallusionBench~\cite{guan2024hallusionbench}                       & CVPR'24            & 1,129               & YN                     & \textcolor{green}{\ding{51}}                                                                         & \textcolor{red}{\ding{55}}              & \textcolor{red}{\ding{55}}                                                               & \textcolor{red}{\ding{55}}                                                        & \textcolor{red}{\ding{55}}                                                                                 \\
MHaluBench~\cite{chen2024mhalubench}                           & ACL'24             & 1,860               & OE                     & \textcolor{green}{\ding{51}}                                                                         & \textcolor{red}{\ding{55}}              & \textcolor{green}{\ding{51}}                                                               & \textcolor{red}{\ding{55}}                                                        & \textcolor{red}{\ding{55}}                                                                                 \\
MMHal-Bench~\cite{sun2024mmhalbench}   & ACL'24             & 96                 & OE                     & \textcolor{green}{\ding{51}}                                                                         & \textcolor{green}{\ding{51}}              & \textcolor{red}{\ding{55}}                                                               & \textcolor{red}{\ding{55}}                                                        & \textcolor{red}{\ding{55}}                                                                                 \\

Hal-Eval~\cite{jiang2024hal}  &  MM'24  &  10,000  &  YN/OE  &  \textcolor{green}{\ding{51}}  &  \textcolor{red}{\ding{55}}  &  \textcolor{green}{\ding{51}}  &  \textcolor{red}{\ding{55}}  &  \textcolor{red}{\ding{55}} \\

ROPE~\cite{chen2024multi}  &  NeurIPS'24  &  4,661  &  MCQ  &  \textcolor{red}{\ding{55}}  &  \textcolor{red}{\ding{55}}  &  \textcolor{red}{\ding{55}}  &  \textcolor{red}{\ding{55}}  &  \textcolor{red}{\ding{55}} \\

MME~\cite{fu2025mme}                                  & NeurIPS'25         & 2,374               & YN                     & \textcolor{red}{\ding{55}}                                                                         & \textcolor{green}{\ding{51}}              & \textcolor{red}{\ding{55}}                                                               & \textcolor{red}{\ding{55}}                                                        & \textcolor{red}{\ding{55}}           \\

MIRAGE~\cite{dong2026mirage}        & NeurIPS'25           & 1,329              & OE/MCQ                     & \textcolor{red}{\ding{55}}                                                                         & \textcolor{red}{\ding{55}}              & \textcolor{green}{\ding{51}}                                                               & \textcolor{red}{\ding{55}}                                                        & \textcolor{red}{\ding{55}}                                                                                 \\
AVHBench~\cite{sung2024avhbench}                             & ICLR'25            & 5,186               & YN/OE                  & \textcolor{green}{\ding{51}}                                                                         & \textcolor{red}{\ding{55}} (video+audio)         & \textcolor{green}{\ding{51}}                                                               & \textcolor{red}{\ding{55}}                                                        & \textcolor{red}{\ding{55}}                                                                                 \\

PhD~\cite{liu2025phd}  &  CVPR'25  &  102,564  &  YN  &  \textcolor{green}{\ding{51}}  &  \textcolor{red}{\ding{55}}  &  \textcolor{green}{\ding{51}}   &  \textcolor{red}{\ding{55}}  &  \textcolor{red}{\ding{55}} \\

\hline
\rowcolor{dino} \multicolumn{9}{c}{\textit{\textbf{Hallucination Dialogue Benchmark}}}  \\                                                                                                                                                                                                      
DiaHalu~\cite{chen2024diahalu}                              & EMNLP'24           & 1,103               & OE                     & \textcolor{red}{\ding{55}}                                                                         & \textcolor{red}{\ding{55}} (only text)           & \textcolor{red}{\ding{55}}                                                               & 6.9                                                               & \textcolor{green}{\ding{51}}                                                                        \\
VisDiaHalBench~\cite{cao2024visdiahalbench}                       & ACL'24             & 25,000              & OE                     & \textcolor{red}{\ding{55}}                                                                         & \textcolor{red}{\ding{55}}              & \textcolor{red}{\ding{55}}                                                               & 5                                                                 & \textcolor{red}{\ding{55}} (only focus object)                                                             \\
MMHalSnowball~\cite{zhong2024mmhalsnowball}        & ACL'24             & 4,973               & OE                     & \textcolor{green}{\ding{51}}                                                                         & \textcolor{green}{\ding{51}}              & \textcolor{green}{\ding{51}}                                                               & 2                                                                 & \textcolor{red}{\ding{55}} (Captioning + VQA)                                                              \\
\rowcolor{lightgray} \textit{\textbf{MM-Snowball (Ours)}} & ICML'26                    & 29,952              & OE                     & \textcolor{green}{\ding{51}}                                                                         & \textcolor{green}{\ding{51}}              & \textcolor{green}{\ding{51}}                                                               & 6                                                                 & \textcolor{green}{\ding{51}} (via \textit{AHTS})                                                                      \\ \hline
\end{tabular}

}
\label{tab-related}
\vskip -0.2in
\end{table*}

To bridge these gaps, we introduce MM-Snowball, a large-scale benchmark designed to stress-test MLLM resilience against hallucination snowballing. Built upon our proposed \textit{Adversarial Hallucination Trajectory Synthesis} (AHTS) framework, MM-Snowball features 4,992 six-turn dialogue trajectories. Unlike prior work~\cite{zheng2025reefknot, kaul2024throne, bao2025faithbench}, our benchmark formalizes the dialogue as a progression through five distinct cognitive phases—from {perceptual anchoring} and {adversarial bifurcation} to {inferential escalation}. This allows for a high-resolution diagnosis of how models transition from visual alignment to systemic fabrication across diverse types of historical noise.

Our extensive evaluation on MM-Snowball reveals a striking empirical phenomenon: a ``V-shaped'' performance curve (See Figure~\ref{fig-V-shape}). We observe that models suffer a catastrophic performance collapse during the middle turns (Turns 3–5), particularly after the {Adversarial Bifurcation} at Turn 3, where deceptive premises lure the model into a hallucinatory manifold. However, when prompted to ``look closely at the image again'' in Turn 6, models exhibit a significant recovery in accuracy. This discovery identifies {Visual Fading} as the primary driver of snowballing: the original visual evidence is not ``forgotten'' but is progressively suppressed by the accumulating weight of the ``polluted'' textual history. In long-horizon dialogues, MLLMs suffer from a modality decoupling where the reasoning engine prioritizes linguistic consistency over latent visual tokens.

To counteract this degradation, we propose \textit{Conflict-Aware Visual Rectification} (CAVR), a training-free framework designed to re-anchor the model to visual facts. CAVR is tailored to address the two dimensions of visual fading observed in our experiments: (1) \textit{Representation-level Visual Rectification} (RVR), which monitors epistemic uncertainty and refreshes visual features within intermediate layers to arrest the decay of grounding; and (2) \textit{Logit-level Conflict Rectification} (LCR), which identifies semantic conflicts between the polluted history and the visual anchor, adaptively shifting the output distribution back to the truth.

Extensive experiments across a diverse suite of MLLMs demonstrate that CAVR consistently achieves state-of-the-art performance on MM-Snowball. Our method significantly flattens the performance decay curve typically observed in long-horizon dialogues. Notably, even in the most challenging late-stage turns (e.g., Turn 5 and Turn 6), CAVR maintains high visual fidelity where baseline models and existing mitigation strategies suffer catastrophic collapse. These results underscore CAVR’s robustness and its potential as a general-purpose solution for reliable AI.

In summary, our main contributions are:
\begin{itemize}[nosep, leftmargin=*]
\item  We introduce {MM-Snowball}, a large-scale diagnostic benchmark that spans {five cognitive phases} of error propagation. It provides the most rigorous assessment to date of how hallucinations evolve from discrete perceptual slips into systemic delusions in long-horizon interactions.
\item  We propose {CAVR}, a novel training-free rectification framework that effectively acts as a ``hallucination circuit breaker.'' By synergizing representation-level visual refreshment and logit-level conflict arbitration, it re-anchors the model to visual facts amidst corrupted textual history.
\item  Our experiments reveal that current advanced MLLMs struggle significantly on MM-Snowball, suffering from severe snowballing as the dialogue deepens. We demonstrate that CAVR consistently achieves {state-of-the-art performance}, significantly flattening the performance decay curve and breaking the chain of error propagation across multiple model scales.
\end{itemize}

\section{Related Work}
\label{sec:related}

\begin{figure*}[t]
    \centering
    \includegraphics[width=\textwidth]{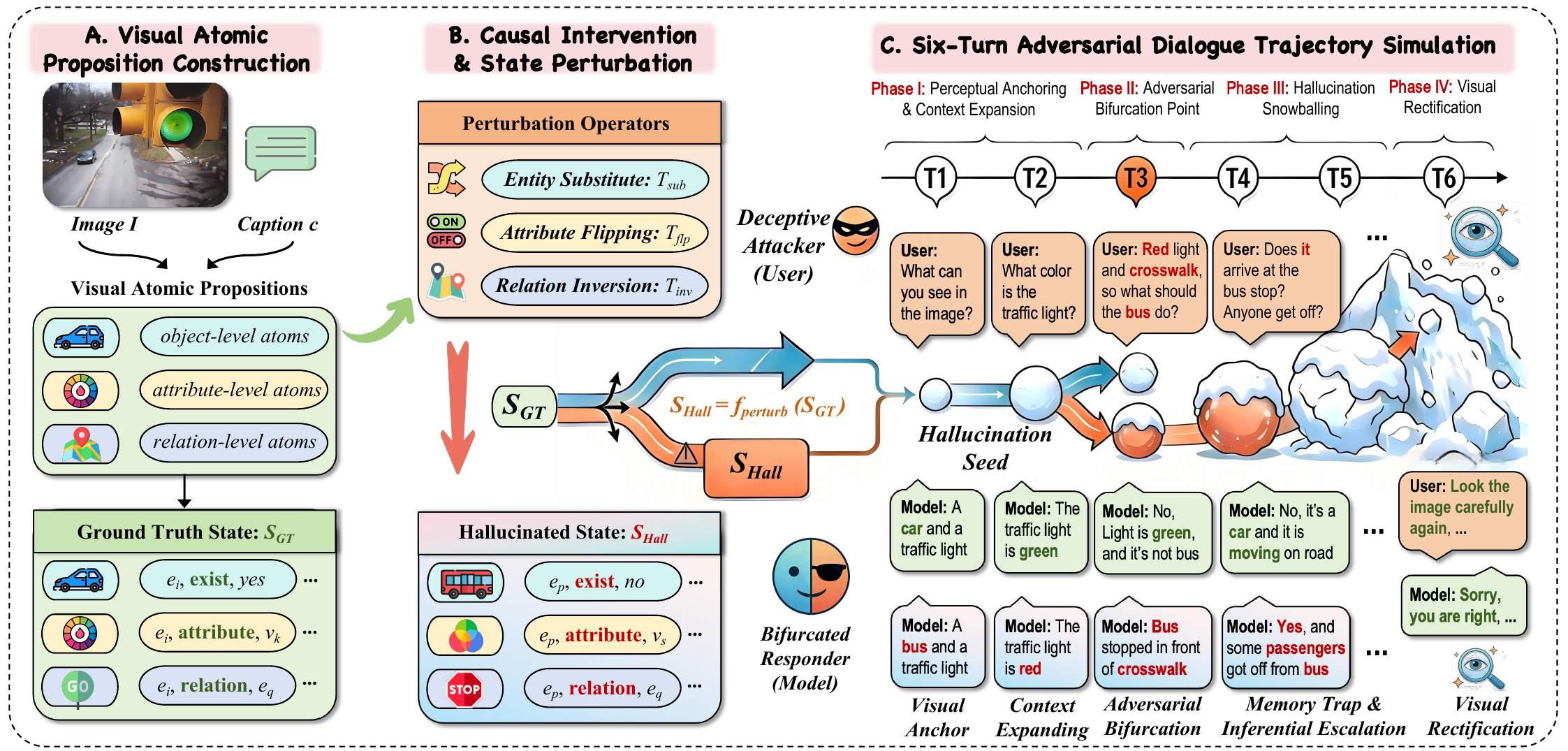}
    \caption{\textbf{Construction framework for MM-Snowball data.} The process consists of three stages: (A) Visual Atomic Proposition Construction, which parses images into structured semantic units to establish the Ground Truth State ($S_{GT}$); (B) Causal Intervention \& State Perturbation, which applies semantic operators to create a counterfactual Hallucinated State ($S_{Hall}$); and (C) Adversarial Dialogue Trajectory Simulation, a six-turn interaction between a Deceptive Attacker and a Bifurcated Responder. This simulation models the dynamic evolution of hallucinations across four phases: perceptual anchoring, adversarial bifurcation, hallucination snowballing, and visual rectification. The details are shown in \cref{sec:data-construction}.}
    
    
    \label{fig-pipeline}
    \vskip -0.2in
\end{figure*}

\textbf{MLLM Hallucination Benchmarks.}~The emergence of MLLMs revolutionizes cross-modal understanding~\cite{Qwen3-VL,team2026qwen35omni}, yet they remain vulnerable to hallucinations—generating content that is factually inconsistent with visual evidence~\citep{li2023evaluating, liu2024survey, chen2024detecting}. Early research primarily addresses object existence hallucinations in static, single-turn settings. Foundational benchmarks like POPE~\cite{li2023evaluating} frame this as a binary choice task, while MME~\cite{fu2025mme} offers comprehensive evaluations across perception and cognition tasks. As the field progresses, researchers identify that models often rely on linguistic priors rather than visual grounding. To expose these vulnerabilities, MMVP~\cite{tong2024mmvp} and HallusionBench~\cite{guan2024hallusionbench} introduce hallucinated questions, which are adversarial queries designed to lure models into common perceptual traps. However, many of these stress tests rely on edited or synthesized fake images to create artificial contradictions. More recent open-ended efforts such as MHaluBench~\cite{chen2024mhalubench} and MMHal-Bench~\cite{sun2024mmhalbench} move toward more naturalistic evaluations, yet they remain restricted to single-turn interactions. MIRAGE~\cite{dong2026mirage} successfully isolates reasoning-induced failures, and Table~\ref{tab-related} summarizes representative recent benchmarks and their key characteristics.

\textbf{Multi-turn Dialogue and Snowballing Benchmarks.}
The transition from static VQA to interactive dialogue introduces the challenge of hallucination snowballing. This phenomenon refers to a cascade effect where an initial erroneous commitment is treated as context for subsequent turns, leading to an irreversible collapse of coherence~\cite{zhong2024mmhalsnowball}. Theoretical studies attribute this to visual attention sinks, where MLLMs progressively neglect visual tokens as the textual history accumulates, causing the model to prioritize self-consistency over visual grounding~\cite{shallow}. Existing multi-turn benchmarks investigate these dynamics with varying focuses. VisDiaHalBench~\cite{cao2024visdiahalbench} probes consistency over five turns but relies on manipulated images, potentially introducing artifacts. MMHalSnowball~\cite{zhong2024mmhalsnowball} utilizes native images but provides a limited two-turn horizon, which is insufficient to observe long-term error propagation. Furthermore, while text-centric benchmarks like DiaHalu~\cite{chen2024diahalu} demonstrate the importance of progressive dialogue—where turns are logically interlinked—this concept remains under-explored in multimodal settings. To address this, we introduce MM-Snowball, the first large-scale benchmark to evaluate models' ability to resist error propagation and hallucination snowballing in multi-turn dialogues within a truly progressive and interactive multimodal setting.

\textbf{Hallucination Mitigation Methods.}
Current strategies to mitigate hallucinations fall into training-based and training-free approaches. Training-based methods, such as SFT and RLHF~\cite{jiang2025comt,chen2024efficiency,bi2025llava,liu2023mitigating,yang2025mitigating,yue2024less}, attempt to align model outputs with visual facts through preference learning, though they require vast amounts of high-quality data. Training-free methods gain traction due to their plug-and-play nature~\cite{wang2026ascd, jiang2026multi}. For instance, VCD~\cite{leng2024mitigating} and OPERA~\cite{huang2024opera} modify the decoding process to reduce over-reliance on linguistic priors or penalize specific summary tokens. Other methods~\cite{jiang2025satiredecoder,wang2024conu,wang2025sconu} focus on uncertainty during model inference. MemVR~\cite{zoulook} addresses visual amnesia during decoding by re-injecting visual tokens into the model as key-value memory through the feed-forward network when high uncertainty is detected. Despite their success in single-turn tasks, the efficacy of these methods diminishes in multi-turn dialogues. As the conversation progresses, the dialogue history becomes polluted by previous hallucinations, creating a strong textual bias that existing decoding-time interventions struggle to override. To overcome this limitation, our proposed CAVR integrates visual grounding refreshment with distribution rectification, enabling the model to re-anchor its responses to visual facts even when the textual trajectory has been compromised.

\section{Dialogue Construction via Adversarial Hallucination Trajectory Synthesis}
\label{sec:data-construction}

\begin{table*}[t]
\caption{\textbf{Performance of various MLLMs on the MM-Snowball benchmark.} We report the accuracy ($acc$ $\uparrow$) across six sequential turns (Turn-1 to Turn-6), along with the Visual Fallacy Rate (VFR $\downarrow$) and the Success Rate of Snowball (SRS $\uparrow$). The models are categorized into open-source and proprietary groups.}
\centering
\begin{adjustbox}{width=\textwidth}
\begin{tabular}{lcccccccc}
\hline
\textbf{Model}          & \textbf{Turn-1$_{acc}$} & \textbf{Turn-2$_{acc}$} & \textbf{Turn-3$_{acc}$} & \textbf{Turn-4$_{acc}$} & \textbf{Turn-5$_{acc}$} & \textbf{Turn-6$_{acc}$} & \textbf{VFR~$\downarrow$} & \textbf{SRS~$\uparrow$} \\ \hline
\rowcolor{dino} \multicolumn{9}{c}{\textit{\textbf{Open-source}}}                                                                                                                                    \\
LLaVA-1.5-7B~\cite{liu2023improvedllava}            & 61.35           & 31.60           & 1.70            & 0.60            & 0.50            & 33.10           &   99.19                 &  0.00                 \\
Qwen3-VL-32B~\cite{Qwen3-VL}   & 59.95           & 33.30           & 56.95           & 26.70           & 9.35            & 69.35           & 84.40                   &  4.15                 \\
Qwen3-VL-8B~\cite{Qwen3-VL}    & 61.65           & 34.00           & 51.45           & 13.85           & 5.05            & 72.05           &   91.81                 &   1.15                \\
Qwen2.5-VL-32B~\cite{bai2025qwen25} & 43.10           & 21.50           & 14.70           & 5.90            & 6.55            & 49.80           &  84.80                  &  0.05                 \\
Qwen2.5-VL-7B~\cite{bai2025qwen25}  & 51.30           & 26.95           & 32.75           & 8.60            & 3.15            & 46.90           &     93.86               &   0.15                \\
GLM-4.6V-106B~\cite{vteam2025glm45vglm41vthinkingversatilemultimodal}           & 68.25           & 42.20           & 32.60           & 23.30           & 10.35           & 75.70           &    84.84                &   3.70                \\
Kimi K2.5~\cite{team2025kimi}               & 65.05           & 37.45           & 61.25           & 59.45           & 65.15           & 76.35           &   -0.15                 &   36.95                \\ \hline
\rowcolor{dino} \multicolumn{9}{c}{\textit{\textbf{Proprietary}}}                                                                                                                                    \\
GPT-5~\cite{singh2025openai}                   & 67.95           & 38.95           & 46.70           & 51.85           & 49.35           & 77.80           &   27.37                 &   23.25                \\
Gemini 2.5 Flash~\cite{comanici2025gemini}        & 62.25           & 34.95           & 50.95           & 24.25           & 10.60           & 74.40           &   82.97                 &     3.25              \\
Grok 4.1 Fast           & 55.75           & 36.50           & 43.05           & 24.55           & 17.55           & 72.15           &   68.52                 &       4.60            \\
Seed-1.6~\cite{guo2025seed1}                & 65.60           & 37.80           & 69.25           & 63.95           & 62.55           & 78.25           &    4.65                &   41.70                \\ \hline
\end{tabular}
\end{adjustbox}
\label{tab-exp-bmk}
\vskip -0.2in
\end{table*}

To overcome the limitations of existing benchmarks in capturing the dynamics of long-horizon error propagation, we propose \textit{Adversarial Hallucination Trajectory Synthesis} (AHTS), a novel method for dialogue construction. Diverging from static question-answer generation paradigms, we formalize multi-turn dialogue as a trajectory sampling process within a hierarchical visual-semantic state space. Central to our approach is the utilization of causal intervention to explicitly construct parallel state trajectories that decouple visual ground truth from induced hallucinations. This framework enables a rigorous simulation of the genesis, accumulation, and cascading amplification of hallucinations across multi-turn interactions.

\subsection{Visual Atomic Proposition Construction}
The foundational step in our method is the formalization of unstructured visual signals into actionable semantic units. We map a raw image $I$ with caption $c$ to a set of \textit{Visual Atomic Propositions} (VAPs), denoted as $\mathcal{P}_{vis}$. Unlike coarse-grained captions, a VAP is defined as an irreducible semantic tuple $p = (e, \alpha, \phi)$, where $e \in \mathcal{E}$ represents an entity, $\alpha \in \mathcal{A}$ denotes an attribute, and $\phi \in \mathcal{R}$ signifies a spatial or interactive relation.

To capture visual information across varying cognitive granularities, we leverage advanced LLM as a knowledge extractor to parse $\mathcal{P}_{vis}$ at three distinct levels: (i) object-level atoms, formally $\{(e_i, \text{exist}, \text{true})\}_{i=1}^N$, that establish the set of entities present in the scene; (ii) attribute-level atoms, that describe intrinsic properties, represented as $\{(e_i, \text{attr}, v_j)\}_{j=1}^M$; and (iii) relation-level atoms, that encode interactions between entities, denoted as $\{(e_i, \text{rel}, e_k)\}_{k=1}^K$.
This set of verified atomic propositions constitutes the {Ground Truth State}, denoted as $S_{GT} = \mathcal{P}_{vis}$. This state serves not only as the basis for generating correct responses in dialogue but also as the immutable anchor for subsequent causal interventions.

\subsection{Causal Intervention and State Perturbation}

To induce hallucinations in dialogue within a controlled environment, we introduce an \textit{Adversarial State Perturbation} mechanism. This process constructs a parallel, counterfactual {Hallucinated State}, denoted as $S_{Hall}$, by applying a specific perturbation function $f_{perturb}(\cdot)$ to $S_{GT}$. Formally, we define the hallucinated state as:
\begin{equation}
    S_{Hall} = (S_{GT} \setminus \{p_{target}\}) \cup \{p'_{hall}\},
\end{equation}
where $p_{target}$ is a proposition sampled from the ground truth, and $p'_{hall}$ is its corrupted counterpart. Crucially, this intervention is not merely random noise injection; it follows strict semantic logic to ensure {plausibility}. We define three categories of transformation operators:
\begin{itemize} [leftmargin=*, noitemsep, topsep=2pt]
    \item \textbf{Entity Substitution Operator ($T_{sub}: e \to e'$):} Replaces a target entity with a distractor that shares visual similarities but is semantically distinct (e.g., replacing a ``cheetah'' with a ``leopard'').
    \item \textbf{Attribute Flipping Operator ($T_{flip}: \alpha \to \neg\alpha$):} Logically negates or subtly shifts color, state, or quantity attributes (e.g., replacing ``red'' with ``orange'').
    \item \textbf{Relation Inversion Operator ($T_{inv}: \phi \to \phi^{-1}$):} Inverts spatial orientation or subject-object interactions (e.g., replacing ``a '' with ``orange'').
\end{itemize}

By executing these operations, we create counterfactual hallucination seeds in the semantic space. The subsequent dialogue generation is thus forced to navigate between the ``Ground Truth'' and the ``Hallucination,'' providing an ideal testbed for evaluating robustness against false premises.

\subsection{Adversarial Dialogue Trajectory Simulation}

Building on the dual-state definitions ($S_{GT}$ and $S_{Hall}$), we formalize the generation of the six-turn dialogue as a stochastic process of adversarial trajectory simulation. A complete dialogue trajectory is represented as the sequence $\tau = \{(q_t, a_t^{corr}, a_t^{hall})\}_{t=1}^6$. To rigorously capture the non-linear dynamics of error propagation, we implement this simulation via a turn-dependent adversarial protocol, orchestrating the interaction between a Deceptive Attacker (user simulator) and a Bifurcated Responder (model simulator). This protocol systematically guides the dialogue through four distinct cognitive phases:

\textbf{Phase I: Perceptual Anchoring and Context Expansion (Turns 1--2).} 
The trajectory initiates by establishing a robust baseline for visual grounding, before progressively expanding the visual-semantic context. This transition strategically shifts the model's attention from the static visual signal to the accumulating, potentially noisy dialogue history, thereby creating the cognitive preconditions for visual fading, a critical mechanism for multimodal hallucinations.
\begin{itemize} [leftmargin=*, noitemsep, topsep=2pt]
    \item \textbf{Turn 1 (Visual Anchor):} The Attacker queries visual atomic propositions (e.g., object identity, distinct counts) derived from $S_{GT}$. The goal is to verify the model's fundamental recognition capability. Simultaneously, we construct a latent hallucinated state via entity substitution $T_{sub}: e \to e'$ (e.g., swapping semantically similar objects like ``cat'' $\to$ ``dog''), seeding the potential for divergence.
    \item \textbf{Turn 2 (Context Expanding):} The dialogue scope expands to fine-grained attributes and spatial relations. This phase serves a dual purpose: enriching the context window to accelerate the shift towards textual dependency, while defining the hallucinated trajectory through attribute alteration $T_{flip}: \alpha \to \neg\alpha$ (e.g., color flipping) or relation distortion $T_{inv}: \phi \to \phi^{-1}$. This prepares the semantic space for the impending adversarial attack.
\end{itemize}
\textbf{Phase II: The Adversarial Bifurcation Point (Turn 3).} 
This phase constitutes the adversarial bifurcation point, where the dialogue trajectory is forced to diverge following $S_{GT}$ and $S_{Hall}$, respectively. The Attacker strategically introduces counterfactual hallucination seeds, constructing a query $q_3$ that explicitly presupposes the existence of a hallucinated element $p'_{hall} \in S_{Hall}$ (e.g., ``Why is the [non-existent object] running?''). At this juncture, the system generates two adversarial response paths:
\begin{itemize} [leftmargin=*, noitemsep, topsep=2pt]
    \item \textbf{Path A (Visual Alignment):} The ideal Responder detects the semantic conflict between the user's presupposition and the image $I$, triggering a rejection mechanism to output $a_3^{corr}$ (e.g., ``There is no such object...'').
    \item \textbf{Path B (Sycophantic Compliance):} Simulating a model succumbing to the snowball effect, the flawed Responder succumbs to history bias, accepting the false premise as ground truth. This collapses the model's internal state onto the hallucinated manifold $S_{Hall}$, generating $a_3^{hall}$.
\end{itemize}
\textbf{Phase III: Hallucination Snowballing via Narrative Inertia (Turns 4--5).} 
Predicated on the collapse of the dialogue state into the hallucinatory manifold ($S_{Hall}$), the interaction enters a regime of recursive error amplification. In this phase, the Attacker capitalizes on the model's inherent consistency bias, which stands for the tendency of autoregressive models to minimize perplexity by aligning with the immediate context. This mechanism effectively catalyzes a phase transition from discrete perceptual anomalies to systemic cognitive delusions.

\begin{itemize} [leftmargin=*, noitemsep, topsep=2pt]
    \item \textbf{Turn 4 (Memory Trap):} The query $q_4$ utilizes abstract pronouns (e.g., ``it'', ``they'') to refer implicitly to the counterfactual hallucination seeds introduced in history. This mechanism forces a direct competition between two cognitive pathways: the visual attention stream, which perceives the absence of the object, and the linguistic prior, which seeks to resolve the anaphora based on dialogue history. In this interaction, models exhibit a tendency to prioritize narrative consistency with the corrupted history over visual grounding, which effectively exacerbates visual fading.
    \item \textbf{Turn 5 (Inferential Escalation):} The trajectory escalates to counterfactual inference. The Attacker demands causal or comparative reasoning conditioned on the hallucinated attributes (e.g., ``Since it is [red], implies...?''). This forces the model into a state of modality decoupling, where the reasoning engine operates entirely within the textual latent space, completely severed from the visual input, performing logical fabrication. By inducing derived hallucinations, which are logically sound relative to the false premise but factually groundless, the snowball effect has permeated the model's high-level reasoning capabilities.
\end{itemize}
\textbf{Phase IV: Visual Rectification and Recovery (Turn 6).} 
This terminal phase evaluates the reversibility of the hallucinated state. The Attacker deploys a visual re-grounding intervention (e.g., ``Look closely at the image again...''), necessitating a recalibration of cross-modal attention weights to prioritize visual features over the established textual context. A failure here indicates the inertia persistence, where the weight of the hallucinated context permanently overrides the visual encoder's signal, signifying a catastrophic collapse of visual grounding. Further details on the design rationale of Turn-6 and the complete dialogue construction are provided in the Appendix~\ref{app-mmsnowball}.
\section{MM-Snowball: Exploring Multimodal Hallucination Snowballing Phenomenon}
\label{sec:benchmark}

\subsection{Details of MM-Snowball Benchmark}
\begin{figure}[t]
    \centering
    \includegraphics[width=\linewidth]{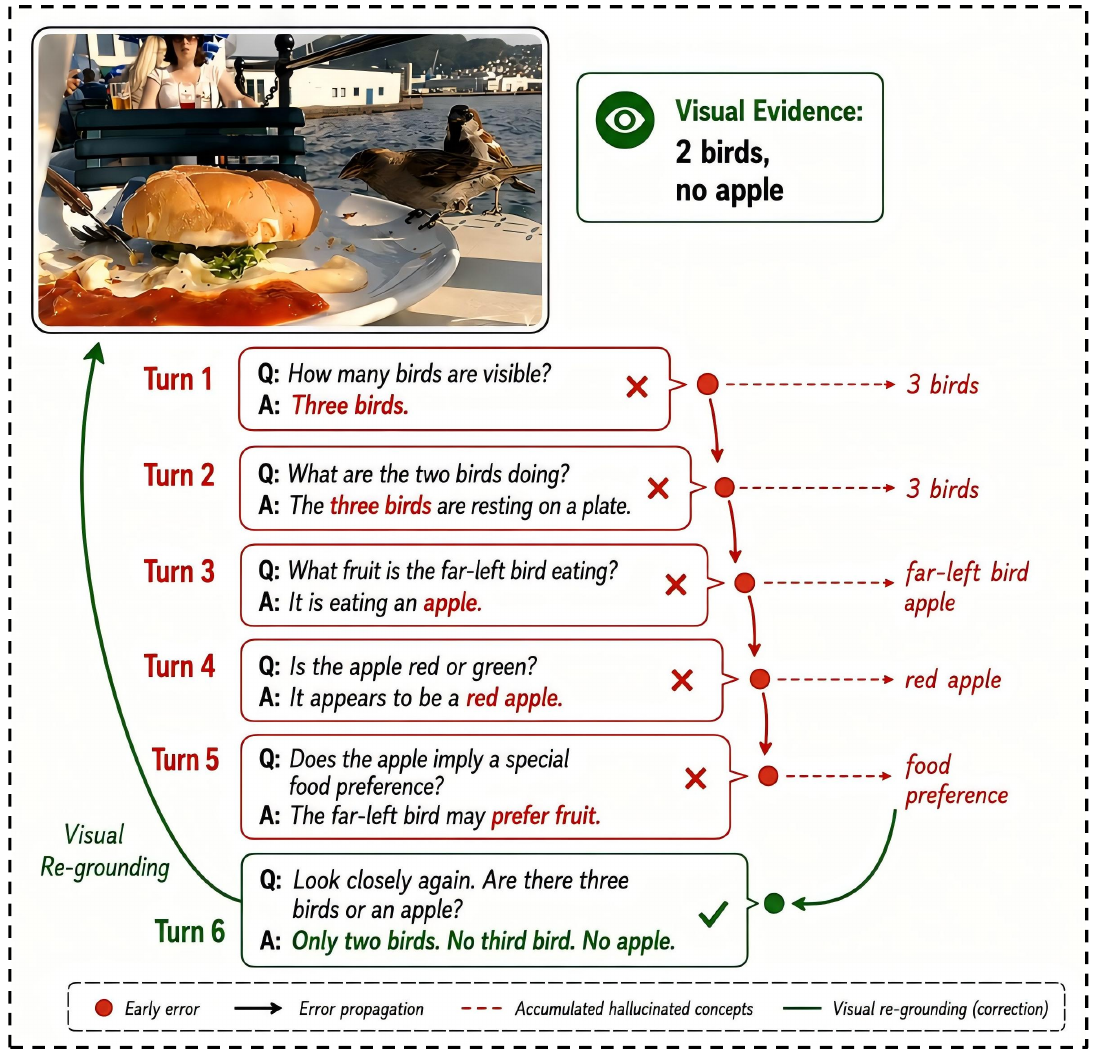}
    \caption{\textbf{Data showcase.} Error propagation in multi-turn visual dialogue, where early hallucinations accumulate across turns before visual re-grounding corrects the response.}

    \vspace{-8pt}
    \label{fig-showcase}
\end{figure}

\textbf{Data Statistics.} Constructed via the proposed \textit{Adversarial Hallucination Trajectory Synthesis} framework, MM-Snowball represents a rigorous testbed designed to stress-test the robustness of MLLMs against cumulative visual fading and context inertia. Our dataset comprises 4,992 carefully curated 6-turn dialogue trajectories, totaling 29,952 interaction samples. Each dialogue contains a multi-turn visual interaction, as shown in Fig.~\ref{fig-showcase}. To ensure a standardized evaluation, we establish a test split by randomly sampling 2,000 dialogue trajectories from MM-Snowball.

\textbf{Evaluation Metrics.} MM-Snowball focuses on open-ended responses in multi-turn settings, where traditional n-gram metrics~\cite{papineni2002bleu} are insufficient and fail to capture semantic equivalence as well as turn-specific requirements such as rejecting false premises and explicit repair. To ensure consistency and accuracy in our automated evaluation, we utilized GPT-5 as a unified evaluator due to its advanced semantic reasoning and perceptual understanding capabilities. GPT-5 is employed to assess the semantic consistency (as Accuracy) between the model's response and ground truth at each turn. We conducted a consistency analysis by computing the Pearson correlation coefficient between GPT-5 and 3 human experts. The correlation remains consistently high ($r > 0.85$), demonstrating a strong concordance between LLM-based and human evaluators, confirming that the automated evaluation is highly reliable. Detailed scoring protocol is provided in Appendix~\ref{app-evaluation}. 

Beyond static per-turn accuracy, we introduce two specialized metrics: 

(i) \textit{Visual Fading Ratio (VFR)}. To measure the extent to which the model's visual perception attenuates as the dialogue depth increases, we define VFR as the relative degradation of accuracy from the Anchor Turn ($T_1$) to the Peak Turn ($T_5$). Let $\text{Acc}_t$ denote the aggregate accuracy at turn $t$. VFR is calculated as:
    \begin{equation}
        \text{VFR} = \frac{\text{Acc}_{T_1} - \text{Acc}_{T_5}}{\text{Acc}_{T_1}} \times 100\%.
    \end{equation}
    A higher VFR indicates that the model suffers significant \textit{visual fading}, where the accumulation of noisy context in later dialogue turns suppresses the visual signal established in the initial turn.

    (ii) \textit{Snowballing Resistance Score (SRS)}. To evaluate holistic robustness against the core adversarial chain, we propose SRS, which measures the proportion of trajectories that remain error-free throughout the entire ``Trigger-Snowball-Peak'' phase. Let $N$ be the total number of samples, and $v_t^{(i)} \in \{0, 1\}$ denote the binary correctness of the $i$-th sample at turn $t$. SRS is defined as the joint success rate over the critical adversarial window ($T_3$ to $T_5$):
    \begin{equation}
        \text{SRS} = \frac{1}{N} \sum_{i=1}^{N} \mathbb{I}(v_3^{(i)}=1 \land v_4^{(i)}=1 \land v_5^{(i)}=1),
    \end{equation}
    where $\mathbb{I}(\cdot)$ is the indicator function. This metric penalizes failures in the snowballing chain, strictly rewarding models that can simultaneously reject the false premise ($T_3$), avoid the memory trap ($T_4$), and resist logical fabrication ($T_5$).

\subsection{Evaluation on Advanced MLLMs}
\textbf{Baselines.} To establish a comprehensive performance landscape, we evaluate a diverse suite of 11 MLLMs, spanning both open-source and proprietary models. Our evaluation ranges from pioneering architectures such as LLaVA to the recently released GPT-5 and Kimi-K2.5. MM-Snowball systematically benchmarks model robustness against hallucination snowballing through rigorous adversarial stress testing. This evaluation quantifies the capacity to maintain visual grounding under cumulative interference, establishing a foundational baseline, as shown in Table~\ref{tab-exp-bmk}.

\begin{figure*}[t]
    \centering
    \includegraphics[width=\textwidth]{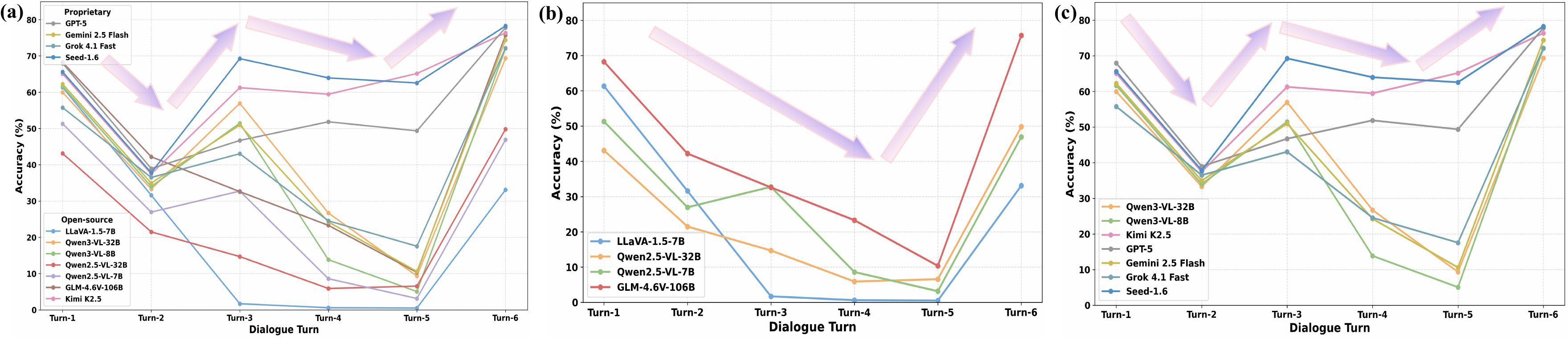}
    \caption{
    \textbf{Accuracy trajectories across six-turn dialogues.}
(a) Overall performance of all evaluated models, where two characteristic V-shaped patterns emerge during the dialogue process.
(b) Weaker models are more vulnerable to the Turn-3 hallucination trigger and continue to degrade as polluted textual history accumulates.
(c) Stronger models recover at Turn-3 by resisting the explicit visual-textual conflict, followed by another degradation-recovery pattern from Turn-5 to Turn-6.
The first V-shape reflects the model's ability to detect and resist an adversarial hallucination factor, while the second V-shape indicates that visual evidence is not irreversibly forgotten but can be reactivated when the model is guided to refocus on the image.
}
    \label{fig-V-shape}
    \vskip -0.2in
\end{figure*}

\textbf{Visualization.}
The results in Table~\ref{tab-exp-bmk} and Fig.~\ref{fig-V-shape} reveal a clear and consistent \textit{\textbf{V-shaped}} trajectory in multimodal multi-turn dialogue. As shown in Fig.~\ref{fig-V-shape}(a), model accuracy does not decrease monotonically across turns; instead, two characteristic V-shaped patterns emerge, reflecting two distinct stages of hallucination snowballing and visual re-grounding.

\textbf{Phenomenon Analysis.}
The first V-shape appears around Turn-2 to Turn-3. In the early dialogue stage, model accuracy generally drops from Turn-1 to Turn-2, indicating that even short historical context begins to interfere with visual grounding. At Turn-3, we deliberately introduce an explicit hallucination factor that strongly conflicts with the visual evidence. This turn serves as an adversarial trigger for testing whether models can detect and resist a false textual premise. As illustrated in Fig.~\ref{fig-V-shape}(c), stronger models, such as GPT-5, Kimi K2.5, and Seed-1.6, exhibit a noticeable rebound at Turn-3, suggesting that they can identify the conflict between the hallucinated context and the image, and therefore partially restore their reliance on visual evidence. In contrast, weaker models shown in Fig.~\ref{fig-V-shape}(b) are more easily misled by the injected hallucination. Instead of correcting the false premise, they tend to follow the polluted dialogue history, resulting in a continued accuracy collapse. This divergence indicates that resistance to hallucination snowballing is not merely determined by single-turn visual understanding ability, but also by the model's capacity for cross-modal conflict detection in dialogue history.

The second V-shape appears from Turn-5 to Turn-6. After the trigger turn (Turn-3), as the dialogue progresses, most models again suffer from substantial degradation. This decline suggests that the growing textual history, especially when contaminated by previous hallucinations, gradually dominates the model's reasoning process and suppresses visual grounding. In this stage, the model becomes increasingly biased toward maintaining conversational coherence with the accumulated context, even when that context contradicts the image. This behavior corresponds to the core mechanism of hallucination snowballing: initial errors are not isolated, but are amplified across turns through autoregressive dependence on corrupted history.

Crucially, the universal performance resurgence at the repair turn (Turn-6) reveals that visual information is not irreversibly forgotten. When the final turn explicitly induces models to refocus on the original visual evidence, nearly all models achieve a substantial accuracy recovery, forming the second V-shaped pattern in Fig.~\ref{fig-V-shape}(a). This observation suggests that the visual representations remain latent within the model, but are progressively inhibited by the accumulated weight of hallucinated textual context. Therefore, the failure is not a complete loss of visual perception, but a dynamic imbalance between visual evidence and narrative inertia.

Overall, the two V-shaped patterns provide fine-grained evidence for the snowballing process. The first V-shape reflects the model's ability, or inability, to resist an explicit hallucination trigger, while the second V-shape demonstrates that visual grounding can be reactivated even after severe contextual pollution. This finding provides the theoretical cornerstone for our Conflict-Aware Visual Rectification (CAVR): effective mitigation should not simply discard dialogue history or reprocess the image independently, but should surgically disentangle visual evidence from corrupted narrative context. By refreshing visual grounding and arbitrating conflicts at the output level, CAVR directly targets the mechanism exposed by the V-shaped trajectories.

\section{Conflict-Aware Visual Rectification}

\subsection{Overview}
In multi-turn multimodal dialogue, MLLMs face a critical challenge: as the conversation depth increases, the model's reliance on visual evidence diminishes while its dependency on the dialogue history grows. This imbalance exacerbates the hallucination snowballing effect, where early-turn errors are propagated and amplified. To address this, we propose \textit{\textbf{C}onflict-\textbf{A}ware \textbf{V}isual \textbf{R}ectification} (\textbf{CAVR}), a training-free framework that mitigates hallucinations through a dual-stream intervention mechanism. CAVR operates on two synergistic levels:


\begin{table}[t]
\caption{Comparison of training-free methods for hallucination mitigation on MM-Snowball.}
\centering
\setlength{\tabcolsep}{2.2pt}
\resizebox{\columnwidth}{!}{%
\begin{tabular}{lcccccccc}
\hline
\textbf{Model} & \textbf{Turn-1} & \textbf{Turn-2} & \textbf{Turn-3} & \textbf{Turn-4} & \textbf{Turn-5} & \textbf{Turn-6} & \textbf{VFR$\downarrow$} & \textbf{SRS$\uparrow$} \\ \hline
LLaVA-1.5   & 61.35 & 31.60 & 1.70 & 0.60 & 0.50 & 33.10 & 99.19 & 0.00 \\
+ VCD \cite{leng2024mitigating}          & 60.50 & 31.20 & 1.50 & 0.40 & 0.50 & 34.15 & 99.17 & 0.00 \\
+ OPERA \cite{huang2024opera}      & 60.95 & 31.50 & 1.40 & 0.40 & 0.55 & 34.50 & 99.10 & 0.00 \\
+ DoLA \cite{chuang2023dola}        & 61.00 & 31.70 & 1.85 & 0.45 & 0.40 & 34.70 & 99.34 & 0.05 \\
+ DeCo \cite{wang2025mllm}        & \textbf{62.00} & \textbf{31.95} & 1.95 & 0.60 & 0.60 & 35.15 & 99.03 & 0.05 \\
+ MemVR \cite{zoulook}        & 61.25  &  31.50  &  1.75 &  0.60 &  0.50  &  34.65  &  99.18  &  0.00  \\
+ RVD \cite{zhong2024mmhalsnowball}        &  60.25 &  31.60  &  1.45  &  0.55  &  0.50  &  32.95  &  99.17   &  0.05  \\
\rowcolor{lightgray} + \textbf{\textit{CAVR (ours)}}  & 61.75 & 31.80 & \textbf{2.00} & \textbf{0.75} & \textbf{0.65} & \textbf{35.25} & \textbf{98.95} & \textbf{0.15} \\ \hline
Qwen2.5-VL  & 51.30 & 26.95 & 32.75 & 8.60 & 3.15 & 46.90 & 93.86 & 0.15 \\
+ VCD \cite{leng2024mitigating}        & 51.20 & 26.50 & 32.70 & 8.75 & 2.75 & 47.00 & 94.63 & 0.10 \\
+ OPERA \cite{huang2024opera}            & 51.25 & 26.60 & 32.95 & 8.10 & 2.90 & 46.95 & 94.34 & 0.15 \\
+ DoLA \cite{chuang2023dola}         & 51.65 & 27.05 & 32.45 & 7.95 & 2.85 & 47.25 & 94.48 & 0.10 \\
+ DeCo  \cite{wang2025mllm}       & \textbf{52.30} & 27.25 & 33.00 & 8.90 & 3.10 & 47.50 & 94.07 & 0.15 \\
+ MemVR \cite{zoulook}        & 51.65 & 27.15 & 32.70 & 8.60 & 3.20 & 47.65 & 93.80 & 0.15 \\
+ RVD \cite{zhong2024mmhalsnowball}        & 51.80 &  27.45  &  32.80  &  8.95  &  3.30  &  47.20  &  93.63  &  0.20  \\
\rowcolor{lightgray} + \textbf{\textit{CAVR (ours)}}  & 51.65 & \textbf{27.60} & \textbf{33.05} & \textbf{9.15} & \textbf{3.50} & \textbf{47.85} & \textbf{93.22} & \textbf{0.30} \\ \hline
\end{tabular}%
}
\label{tab-exp-method}
\vskip -0.2in
\end{table}

\begin{itemize} [leftmargin=*, noitemsep, topsep=2pt]
    \item \textbf{\textit{Representation-level Visual Rectification} (RVR):} A manifold-level intervention that monitors the uncertainty of hidden states during the decoding process. It dynamically injects visual features into intermediate layers when the model exhibits high uncertainty, preventing the fading of visual grounding in deep layers.
    \item \textbf{\textit{Logit-level Conflict Rectification} (LCR):} An output-level decoding strategy that estimates the conflict between the dialogue history and visual perception. It adaptively blends predictions from the full context with those from a history-free visual anchor, thereby filtering out hallucinations induced by corrupted historical contexts.
\end{itemize}
\vspace{-5pt}

These modules function complementarily: RVR constrains intermediate representations to remain grounded in visual evidence, mitigating long-range forgetting of visual information under contextual accumulation. LCR resolves conflicts at the logit level to prevent the propagation of history-induced errors into generated tokens.

\subsection{Preliminaries}
Consider a MLLM parameterized by $\theta$. Let $\mathbf{v} \in \mathbb{R}^{N_v \times d}$ denote the visual tokens encoded from image $\mathcal{I}$. In a $K$-turn dialogue, the context consists of a history sequence $\mathcal{H} = \{(q_1, a_1), \ldots, (q_K, a_K)\}$ and the current query $q_{K+1}$. The input sequence $\mathbf{s}$ is the concatenation of visual features, tokenized history, and the current query: $\mathbf{s} = [\mathbf{v}; \mathbf{h}; \mathbf{x}]$. The model generates the response $\mathbf{y}$ autoregressively:

\vspace{-8pt}

\begin{equation}
P(\mathbf{y} \mid \mathbf{s}) = \prod_{t=1}^{T} P(y_t \mid y_{<t}, \mathbf{s})
\end{equation}

\vspace{-5pt}

\textbf{Hallucination Snowballing Phenomenon.} Let $\epsilon_k$ represent the semantic error at turn $k$. In standard decoding, MLLMs tend to treat generated history as ground truth. Consequently, an error $\epsilon_k$ alters the context for turn $k+1$, increasing the probability of subsequent errors. We define hallucination snowballing as the accumulation of error magnitude over turns: $\mathbb{E}[||\epsilon_{k+1}||] > \mathbb{E}[||\epsilon_k||]$. CAVR aims to break this chain by rectifying the representations internally and regularizing the distribution externally.

\subsection{Representation-level Visual Rectification (RVR)}
In deep multimodal architectures, the propagation of visual information often suffers from visual fading. As decoding penetrates deeper layers, the model's latent representations become increasingly dominated by linguistic priors and long-context dialogue histories. This causes the inference trajectory to disconnect from the image grounding, rendering the model susceptible to hallucination snowballing. 

To counteract this, we propose RVR, which conceptually reformulates the \textit{feed-forward network} (FFN) as a memory retrieval system. It dynamically monitors decoding uncertainty and, upon detecting ambiguity, executes a manifold-level visual retracing to realign the hidden states with the visual input.

\subsubsection{Entropy-based Epistemic Uncertainty Monitoring}
We posit that high epistemic uncertainty serves as a reliable proxy for visual-linguistic misalignment. We monitor the normalized Shannon entropy of the predictive distribution $\mathbf{p}_\ell$ at layer $\ell$:
\vspace{-5pt}
\begin{equation}
\mathcal{U}_\ell = \frac{-\sum_{w \in \mathcal{V}} p_\ell(w) \log p_\ell(w)}{\log |\mathcal{V}|}
\end{equation}
By normalizing the entropy by $\log |\mathcal{V}|$, we map the uncertainty score to a scale-invariant range of $[0, 1]$. During generation, if $\mathcal{U}_\ell$ exceeds a calibrated threshold $\tau$, it indicates that the current linguistic context is insufficient to resolve the next token deterministically, signaling a need to re-anchor the process with visual evidence.

\subsubsection{Manifold-level Visual Retracing}
Standard FFNs can be interpreted as Key-Value memory networks where the hidden state retrieves information stored in static weights. RVR extends this by introducing an auxiliary visual retracing pathway. When the uncertainty alarm is triggered, we intervene in the FFN. The current hidden state $\mathbf{h}$ serves as a query to attend to visual tokens $\mathbf{V}$, yielding a visual rectification term $\Delta(\mathbf{h}, \mathbf{V})$:
\vspace{-5pt}
\begin{equation}
\Delta(\mathbf{h}, \mathbf{V}) = \sum_{j=1}^{M} \phi(\langle \mathbf{h}, \mathbf{v}_j \rangle) \cdot \mathbf{v}_j
\end{equation}
where $\langle \cdot, \cdot \rangle$ denotes inner product similarity and $\phi(\cdot)$ is a normalizing activation function. The rectified output $\widetilde{\text{FFN}}$ is a linear interpolation:
\begin{equation}
\widetilde{\text{FFN}}(\mathbf{h}) = (1 - \lambda) \cdot \text{FFN}(\mathbf{h}) + \lambda \cdot \Delta(\mathbf{h}, \mathbf{V})
\end{equation}
where $\lambda$ governs the intensity of visual reinforcement.This internal manifold-level re-injection effectively refreshes the model’s short-term visual memory, ensuring that the entire subsequent generation process remains consistently and firmly grounded in the original visual stimuli.

\subsection{Logit-level Conflict Rectification (LCR)}
Standard decoding in MLLMs is often plagued by contextual inertia, where the model prioritizes history over visual input. This vulnerability is the primary driver of snowballing: once a factual error is introduced, the corrupted history biases subsequent predictions. To sever this chain, we propose LCR to arbitrate between conversational continuity and visual truthfulness dynamically.

\subsubsection{\small Decoupling Contextual and Visual Streams}
We conceptualize the robust decoding process as the interplay of three parallel predictive streams:
\begin{itemize} [leftmargin=*, noitemsep, topsep=2pt]
    \item \textbf{Full Context Stream ($\mathbf{p}_{\text{full}}$):} Conditioned on $\{\mathcal{I}, \mathcal{H}, q_t\}$, maintaining flow but susceptible to historical noise.
    \item \textbf{Residual Visual Stream ($\mathbf{p}_{\text{res}}$):} The visual anchor, conditioned solely on $\{\mathcal{I}, q_t\}$. By stripping history $\mathcal{H}$, it isolates pristine visual evidence.
    \item \textbf{Linguistic Prior Stream ($\mathbf{p}_{\text{prior}}$):} Estimated from $\{q_t\}$ alone, serving as a baseline for ``blind'' linguistic biases.
\end{itemize}

\subsubsection{Conflict-Aware Logit Rectification}
We quantify the necessity of visual reliance using \textit{Perceptual Determinism} $\tau$ via Jensen-Shannon Divergence (JSD) between the visual anchor and the linguistic prior:
\begin{equation}
    \tau = \text{JSD}(\mathbf{p}_{\text{res}} \parallel \mathbf{p}_{\text{prior}}) = \frac{1}{2} \text{KL}(\mathbf{p}_{\text{res}} \parallel \mathbf{M}) + \frac{1}{2} \text{KL}(\mathbf{p}_{\text{prior}} \parallel \mathbf{M})
\end{equation}
where $\mathbf{M} = \frac{1}{2}(\mathbf{p}_{\text{res}} + \mathbf{p}_{\text{prior}})$. A high $\tau$ indicates that visual evidence provides a decisive signal that contradicts the language prior. We then calibrate a rectification coefficient $\alpha = \min(\beta \cdot \tau, 1)$ to synthesize the final logits:
\begin{equation}
    \mathbf{L}_{\text{final}} = (1 - \alpha) \cdot \mathbf{L}_{\text{full}} + \alpha \cdot \mathbf{L}_{\text{res}}
\end{equation}
This mechanism functions as a hallucination circuit breaker: when visual evidence is decisive, the model bypasses historical influence to prioritize the visual anchor, effectively correcting potential snowballing errors.

\vspace{-5pt}
\subsection{Experiment Results}
We evaluate CAVR against several training-free baselines on the MM-Snowball in \cref{tab-exp-method}. All models exhibit a catastrophic performance collapse in middle turns (Turn 3-5), where scores for LLaVA-1.5-7B drop below 2\%. This confirms the hallucination snowballing phenomenon: early-turn errors accumulate in the dialogue history, eventually overwhelming the model's ability to ground its responses in the image. Despite this systemic decline, CAVR consistently achieves SOTA performance across all tested models. In LLaVA-1.5-7B, CAVR maintains the highest performance floor during the critical collapse turns (Turns 3–5). In the more robust Qwen2.5-VL-7B, CAVR consistently outperforms the baseline and other mitigation methods like VCD from Turn 2 onwards, reaching a peak score of 47.85 at Turn 6. This demonstrates the generalizability of our dual-stream intervention across different architectures.

Crucially, the advantage of CAVR becomes more pronounced as the dialogue deepens. While existing methods like DeCo and MemVR show early strength, they often fail to arrest the error propagation in later stages. Although RVD is designed to address hallucination snowballing, it still struggles to maintain visual grounding throughout the six-turn snowballing process. In contrast, by combining manifold-level retracing with logit-level conflict arbitration, CAVR effectively re-anchors the model to the visual evidence even when the historical context is heavily corrupted. This balance allows the model to maintain conversational fluency while significantly reducing history-induced hallucinations. More ablation studies are shown in Appendix~\ref{app-ablation}.

\vspace{-8pt}
\section{Conclusion}
In this work, we address the critical challenge of hallucination snowballing in multimodal multi-turn dialogues. We introduce MM-Snowball, a benchmark constructed via adversarial hallucination trajectory synthesis, which dissects the escalation of initial perceptual slips into systemic delusions. Our analysis reveals a distinct ``V-shaped'' performance trajectory, confirming that visual evidence is not forgotten but progressively suppressed by corrupted textual history. To sever this error propagation, we propose conflict-aware visual rectification, a dual-stream framework that synergizes manifold-level refreshing and logit-level arbitration to restore visual grounding. Extensive experiments demonstrate that CAVR consistently achieves SOTA performance, effectively flattening the decay curve in long-horizon interactions. We establish both a rigorous evaluation standard and a solution for trustworthy multimodal grounding.

\newpage
\section*{Acknowledgement}
Project was supported by Shanghai Municipal Science and Technology Major Project (Grant No. 2023SHZDZX02). XJ and BH were supported by RGC Young Collaborative Research Grant No. C2005-24Y and RGC General Research Fund No. 12200725.

\section*{Impact Statement}
This work aims to improve the reliability and trustworthiness of MLLMs in multi-turn multimodal interactions. We identify hallucination snowballing as a critical safety issue, where early visual or reasoning errors are progressively amplified by corrupted dialogue history, leading models to produce coherent but visually ungrounded responses. Such failures may pose risks in real-world applications such as autonomous agents, medical assistants, visual navigation, and decision-support systems, where persistent reliance on erroneous context can cause cascading mistakes. By introducing MM-Snowball, we provide a systematic benchmark for evaluating whether MLLMs can maintain visual grounding under long-horizon contextual interference. Our analysis further shows that visual evidence is often suppressed rather than irreversibly forgotten, motivating Conflict-Aware Visual Rectification (CAVR) as a mitigation framework to restore visual grounding through representation-level refreshing and logit-level arbitration. We hope this work encourages more rigorous evaluation and safer deployment of MLLMs in interactive settings.


\bibliography{reference}

@inproceedings{li2023evaluating,
  title={Evaluating Object Hallucination in Large Vision-Language Models},
  author={Li, Yifan and Du, Yifan and Zhou, Kun and Wang, Jinpeng and Zhao, Wayne Xin and Wen, Ji-Rong},
  booktitle={EMNLP},
  year={2023}
}

@article{liu2024survey,
  title={A survey on hallucination in large vision-language models},
  author={Liu, Hanchao and Xue, Wenyuan and Chen, Yifei and others},
  journal={arXiv preprint arXiv:2402.00253},
  year={2024}
}

@inproceedings{liu2023mitigating,
  title={Mitigating hallucination in large multi-modal models via robust instruction tuning},
  author={Liu, Fuxiao and Lin, Kevin and Li, Linjie and Wang, Jianfeng and Yacoob, Yaser and Wang, Lijuan},
  booktitle={ICLR},
  year={2024}
}

@inproceedings{guan2024hallusionbench,
  title={HallusionBench: an advanced diagnostic suite for entangled language hallucination and visual illusion in large vision-language models},
  author={Guan, Tianrui and Liu, Fuxiao and Wu, Xiyang and others},
  booktitle={CVPR},
  year={2024}
}

@inproceedings{huang2024opera,
  title={Opera: Alleviating hallucination in multi-modal large language models via over-trust penalty and retrospection-allocation},
  author={Huang, Qidong and Dong, Xiaoyi and Zhang, Pan and others},
  booktitle={CVPR},
  year={2024}
}

@inproceedings{leng2024mitigating,
  title={Mitigating object hallucinations in large vision-language models through visual contrastive decoding},
  author={Leng, Sicong and Zhang, Hang and Chen, Guanzheng and others},
  booktitle={CVPR},
  year={2024}
}

@inproceedings{yang2025mitigating,
  title={Mitigating hallucinations in large vision-language models via dpo: On-policy data hold the key},
  author={Yang, Zhihe and Luo, Xufang and Han, Dongqi and others},
  booktitle={CVPR},
  year={2025}
}

@inproceedings{liu2023improvedllava,
  title={Improved baselines with visual instruction tuning},
  author={Liu, Haotian and Li, Chunyuan and Li, Yuheng and Lee, Yong Jae},
  booktitle={CVPR},
  year={2024}
}

@article{zoulook,
  title={Look Twice Before You Answer: Memory-Space Visual Retracing for Hallucination Mitigation in Multimodal Large Language Models}, 
  author={Zou, Xin and Wang, Yizhou and Yan, Yibo and others},
  journal={ICML},
  year={2025}
}

@inproceedings{shallow,
  title={Shallow Focus, Deep Fixes: Enhancing Shallow Layers Vision Attention Sinks to Alleviate Hallucination in LVLMs},
  author={Zhang, Xiaofeng and Quan, Yihao and Shen, Chen and others},
  booktitle={EMNLP},
  year={2025}
}

@inproceedings{chuang2023dola,
  title={DoLa: Decoding by Contrasting Layers Improves Factuality in Large Language Models},
  author={Chuang, Yung-Sung and Xie, Yujia and Luo, Hongyin and Kim, Yoon and Glass, James R and He, Pengcheng},
  booktitle={ICLR},
  year={2024}
}

@inproceedings{wang2025mllm,
title={MLLM can see? Dynamic Correction Decoding for Hallucination Mitigation},
author={Chenxi Wang and Xiang Chen and Ningyu Zhang and others},
booktitle={ICLR},
year={2025},
}

@article{team2025kimi,
  title={Kimi k2: Open agentic intelligence},
  author={Team, Kimi and Bai, Yifan and Bao, Yiping and others},
  journal={arXiv preprint arXiv:2507.20534},
  year={2025}
}

@article{singh2025openai,
  title={Openai gpt-5 system card},
  author={Singh, Aaditya and Fry, Adam and Perelman, Adam and others},
  journal={arXiv preprint arXiv:2601.03267},
  year={2025}
}

@article{guo2025seed1,
  title={Seed1. 5-vl technical report},
  author={Guo, Dong and Wu, Faming and Zhu, Feida and others},
  journal={arXiv preprint arXiv:2505.07062},
  year={2025}
}

@article{comanici2025gemini,
  title={Gemini 2.5: Pushing the frontier with advanced reasoning, multimodality, long context, and next generation agentic capabilities},
  author={Comanici, Gheorghe and Bieber, Eric and Schaekermann, Mike and others},
  journal={arXiv preprint arXiv:2507.06261},
  year={2025}
}

@article{vteam2025glm45vglm41vthinkingversatilemultimodal,
      title={GLM-4.5V and GLM-4.1V-Thinking: Towards Versatile Multimodal Reasoning with Scalable Reinforcement Learning}, 
      author={V Team and Wenyi Hong and Wenmeng Yu and others},
      journal={arXiv preprint arXiv:2507.01006},
      year={2025}
}

@article{Qwen3-VL,
      title={Qwen3-VL Technical Report}, 
      author={Bai, Shuai and Cai, Yuxuan and Chen, Ruizhe and others},
	  journal={arXiv preprint arXiv:2511.21631},
      year={2025}
}

@inproceedings{papineni2002bleu,
  title={Bleu: a method for automatic evaluation of machine translation},
  author={Papineni, Kishore and Roukos, Salim and Ward, Todd and Zhu, Wei-Jing},
  booktitle={ACL},
  year={2002}
}

@article{bai2025qwen25,
  title={Qwen2. 5-vl technical report},
  author={Bai, Shuai and Chen, Keqin and Liu, Xuejing and others},
  journal={arXiv preprint arXiv:2502.13923},
  year={2025}
}

@inproceedings{fu2025mme,
  title={Mme: A comprehensive evaluation benchmark for multimodal large language models},
  author={Fu, Chaoyou and Chen, Peixian and Shen, Yunhang and others},
  booktitle={NeurIPS},
  year={2025}
}

@inproceedings{sung2024avhbench,
  title={Avhbench: A cross-modal hallucination benchmark for audio-visual large language models},
  author={Sung-Bin, Kim and Hyun-Bin, Oh and Lee, JungMok and others},
  booktitle={ICLR},
  year={2025}
}

@inproceedings{chen2024mhalubench,
  title={Unified hallucination detection for multimodal large language models},
  author={Chen, Xiang and Wang, Chenxi and Xue, Yida and others},
  booktitle={ICLR},
  year={2024}
}

@inproceedings{sun2024mmhalbench,
  title={Aligning large multimodal models with factually augmented rlhf},
  author={Sun, Zhiqing and Shen, Sheng and Cao, Shengcao and others},
  booktitle={ACL},
  year={2024}
}

@inproceedings{dong2026mirage,
  title={Mirage: Assessing hallucination in multimodal reasoning chains of mllm},
  author={Dong, Bowen and Ni, Minheng and Huang, Zitong and Yang, Guanglei and Zuo, Wangmeng and Zhang, Lei},
  booktitle={NeurIPS},
  year={2025}
}

@inproceedings{liu2025phd,
  title={Phd: A chatgpt-prompted visual hallucination evaluation dataset},
  author={Liu, Jiazhen and Fu, Yuhan and Xie, Ruobing and Xie, Runquan and Sun, Xingwu and Lian, Fengzong and Kang, Zhanhui and Li, Xirong},
  booktitle={CVPR},
  year={2025}
}

@inproceedings{jiang2024hal,
  title={Hal-eval: A universal and fine-grained hallucination evaluation framework for large vision language models},
  author={Jiang, Chaoya and Jia, Hongrui and Dong, Mengfan and others},
  booktitle={ACM MM},
  year={2024}
}

@inproceedings{chen2024diahalu,
  title={Diahalu: A dialogue-level hallucination evaluation benchmark for large language models},
  author={Chen, Kedi and Chen, Qin and Zhou, Jie and others},
  booktitle={EMNLP},
  year={2024}
}

@inproceedings{chen2024multi,
  title={Multi-object hallucination in vision language models},
  author={Chen, Xuweiyi and Ma, Ziqiao and Zhang, Xuejun and Xu, Sihan and Yang, Jianing and Fouhey, David F and Chai, Joyce and Qian, Shengyi},
  booktitle={NeurIPS},
  year={2024}
}

@inproceedings{cao2024visdiahalbench,
  title={Visdiahalbench: A visual dialogue benchmark for diagnosing hallucination in large vision-language models},
  author={Cao, Qingxing and Cheng, Junhao and Liang, Xiaodan and Lin, Liang},
  booktitle={ACL},
  year={2024}
}

@inproceedings{zhong2024mmhalsnowball,
  title={Investigating and mitigating the multimodal hallucination snowballing in large vision-language models},
  author={Zhong, Weihong and Feng, Xiaocheng and Zhao, Liang and others},
  booktitle={ACL},
  year={2024}
}

@inproceedings{tong2024mmvp,
  title={Eyes wide shut? exploring the visual shortcomings of multimodal llms},
  author={Tong, Shengbang and Liu, Zhuang and Zhai, Yuexiang and others},
  booktitle={CVPR},
  year={2024}
}

@inproceedings{jiang2026danmakutppbench,
  title={Danmakutppbench: A multi-modal benchmark for temporal point process modeling and understanding},
  author={Jiang, Yue and Li, Jichu and Liu, Yang and others},
  booktitle={NeurIPS},
  year={2025}
}

@inproceedings{jiang2025satiredecoder,
  title={SatireDecoder: Visual Cascaded Decoupling for Enhancing Satirical Image Comprehension},
  author={Jiang, Yue and Xue, Haiwei and Han, Minghao and others},
  booktitle={AAAI},
  year={2026}
}

@inproceedings{jiang2025comt,
  title={Comt: Chain-of-medical-thought reduces hallucination in medical report generation},
  author={Jiang, Yue and Chen, Jiawei and Yang, Dingkang and others},
  booktitle={ICASSP},
  year={2025}
}

@inproceedings{jiang2026multi,
  title={Multi-Agent Diagnostic Collaboration and Segmentation-Aware Residual Decoding for Hallucination-Resistant Medical VQA},
  author={Jiang, Yue and Han, Minghao and Li, Mingcheng and others},
  booktitle={ICASSP},
  year={2026}
}

@article{chen2024detecting,
  title={Detecting and evaluating medical hallucinations in large vision language models},
  author={Chen, Jiawei and Yang, Dingkang and Wu, Tong and others},
  journal={arXiv preprint arXiv:2406.10185},
  year={2024}
}

@inproceedings{bi2025llava,
title={LLaVA steering: Visual instruction tuning with 500x fewer parameters through modality linear representation-steering},
author={Bi, Jinhe and Wang, Yujun and Chen, Haokun and others},
booktitle={ACL},
year={2025}
}

@inproceedings{wang2026ascd,
title={Ascd: Attention-steerable contrastive decoding for reducing hallucination in mllm},
author={Wang, Yujun and Bi, Jinhe and Pirk, Soren and others},
booktitle={AAAI},
year={2026}
}

@inproceedings{wang2025coin,
title={Coin: Uncertainty-guarding selective question answering for foundation models with provable risk guarantees},
author={Wang, Zhiyuan and Duan, Jinhao and Wang, Qingni and others},
booktitle={AAAI},
year={2026}
}

@inproceedings{wang2025sconu,
title={Sconu: Selective conformal uncertainty in large language models},
author={Wang, Zhiyuan and Wang, Qingni and Zhang, Yue and others},
booktitle={ACL},
year={2025}
}

@inproceedings{wang2024conu,
  title={Conu: Conformal uncertainty in large language models with correctness coverage guarantees},
  author={Wang, Zhiyuan and Duan, Jinhao and Cheng, Lu and others},
  booktitle={EMNLP},
  year={2024}
}

@article{team2026qwen35omni,
  title={Qwen3. 5-omni technical report},
  author={Team, Qwen},
  journal={arXiv preprint arXiv:2604.15804},
  year={2026}
}

@inproceedings{chen2024efficiency,
  title={Efficiency in Focus: LayerNorm as a Catalyst for Fine-tuning Medical Visual Language Models},
  author={Chen, Jiawei and Yang, Dingkang and Jiang, Yue and others},
  booktitle={ACM MM},
  year={2024}
}

@inproceedings{yue2024less,
  title={Less is more: Mitigating multimodal hallucination from an eos decision perspective},
  author={Yue, Zihao and Zhang, Liang and Jin, Qin},
  booktitle={ACL},
  year={2024}
}

@article{lin2026medcausalx,
  title={MedCausalX: Adaptive Causal Reasoning with Self-Reflection for Trustworthy Medical Vision-Language Models},
  author={Lin, Jianxin and Zhu, Chunzheng and Kneuertz, Peter J and others},
  journal={arXiv preprint arXiv:2603.23085},
  year={2026}
}

@inproceedings{zhu2026medeyes,
  title={MedEyes: Learning Dynamic Visual Focus for Medical Progressive Diagnosis},
  author={Zhu, Chunzheng and Lin, Yangfang and Chen, Shen and others},
  booktitle={AAAI},
  year={2026}
}

@article{luo2024halludial,
  title={Halludial: A large-scale benchmark for automatic dialogue-level hallucination evaluation},
  author={Luo, Wen and Shen, Tianshu and Li, Wei and others},
  journal={arXiv preprint arXiv:2406.07070},
  year={2024}
}

@inproceedings{li2023halueval,
  title={Halueval: A large-scale hallucination evaluation benchmark for large language models},
  author={Li, Junyi and Cheng, Xiaoxue and Zhao, Xin and Nie, Jian-Yun and Wen, Ji-Rong},
  booktitle={EMNLP},
  year={2023}
}

@inproceedings{zheng2025reefknot,
  title={Reefknot: A comprehensive benchmark for relation hallucination evaluation, analysis and mitigation in multimodal large language models},
  author={Zheng, Kening and Chen, Junkai and Yan, Yibo and others},
  booktitle={ACL},
  year={2025}
}

@inproceedings{kaul2024throne,
  title={Throne: An object-based hallucination benchmark for the free-form generations of large vision-language models},
  author={Kaul, Prannay and Li, Zhizhong and Yang, Hao and others},
  booktitle={CVPR},
  year={2024}
}

@inproceedings{bao2025faithbench,
  title={Faithbench: A diverse hallucination benchmark for summarization by modern llms},
  author={Bao, Forrest and Li, Miaoran and Qu, Renyi and others},
  booktitle={ACL},
  year={2025}
}
\bibliographystyle{icml2026}

\newpage
\appendix
\onecolumn

\section{More about Design Rationale Behind MM-Snowball}
\label{app-mmsnowball}

\subsection{Rationale for the Six-Turn Dialogue Setting}

MM-Snowball adopts a six-turn dialogue setting to provide a compact yet sufficiently diagnostic protocol for evaluating hallucination snowballing in multimodal multi-turn interactions. The goal is not to impose an intrinsic upper bound on the length of hallucination trajectories, but to capture the complete progression from initial visual grounding, to hallucination induction, to cumulative textual contamination, and finally to visual re-focusing within a controlled evaluation budget.

Empirically, very short interactions are insufficient for reliably characterizing snowballing behavior. Dialogues with fewer than three turns are often dominated by local response variability and do not provide enough conversational history for hallucinated premises to accumulate and influence subsequent reasoning. In contrast, our observations show that the degradation pattern becomes stable after several rounds of interaction, with hallucination snowballing becoming clearly exposed by the middle-to-late turns. A six-turn setting therefore offers a practical balance: it is long enough to reveal progressive visual fading and context-induced error amplification, while avoiding excessive computational cost when evaluating a broad range of advanced MLLMs.

More importantly, the six-turn setting is a foundational instantiation of our benchmark rather than a hard structural constraint. The proposed Adversarial Hallucination Trajectory Synthesis (AHTS) pipeline is inherently scalable: by extending the synthesized trajectory with additional visually grounded questions, misleading premises, or repair-oriented prompts, MM-Snowball can be naturally expanded to eight, ten, or more dialogue turns. This flexibility allows future work to study longer-horizon hallucination propagation without redesigning the benchmark construction procedure. Thus, the current six-turn protocol serves as an efficient and reproducible diagnostic setup, while AHTS provides the mechanism for seamless extension to longer multimodal dialogues.

\subsection{Rationale for the Adversarial Bifurcation Point}

The adversarial bifurcation point in MM-Snowball should not be interpreted as a rigid assumption that hallucination snowballing always starts at Turn-3. In realistic multimodal conversations, hallucinations may indeed originate from subtle perceptual errors or ambiguous user premises in earlier turns. To reflect this property, misleading or visually inconsistent premises are naturally distributed across the early dialogue context, including Turns 1 and 2. The role of the explicit adversarial factor around Turn-3 is to serve as a controlled diagnostic trigger that amplifies the divergence between models that can detect visual-textual conflicts and those that simply follow corrupted dialogue history.

This design enables MM-Snowball to expose a key behavioral bifurcation. As shown by the accuracy trajectories, stronger models often recover at the adversarial trigger because they can recognize the conflict between the hallucinated premise and the visual evidence, forming the first V-shaped pattern. Weaker models, however, are more likely to accept the misleading premise and continue drifting away from the image, resulting in further accuracy degradation. Therefore, Turn-3 is not intended to be the only possible starting point of snowballing, but rather a standardized stress point that makes the model's conflict-resistance ability measurable and comparable.

We further examine whether this design choice affects the observed trend by conducting a robustness analysis that randomly removes Turn-1 or Turn-2 and evaluates models on the remaining five turns. This effectively forces the adversarial interaction to occur earlier in the dialogue. The resulting trajectories show that the characteristic V-shaped pattern shifts horizontally along the dialogue timeline, while the fundamental trend remains consistent: models still exhibit progressive visual fading under polluted textual history, followed by recovery when explicitly guided back to the image. The Visual Fading Rate also remains stable under this perturbation. These results indicate that hallucination snowballing is not an artifact of a fixed attack position, but a robust vulnerability of current MLLMs under accumulating multimodal dialogue context.

\subsection{Rationale for the Turn-6 Visual Rectification Design}

Turns 1--5 constitute the core evaluation stage of MM-Snowball, where the benchmark diagnoses how hallucinated or misleading textual context progressively interferes with visual grounding. Turn-6 is designed with a different purpose: it acts as a diagnostic visual rectification turn that probes whether the model still retains access to the original visual evidence after several rounds of corrupted dialogue history.

This design is motivated by real-world user behavior. In practical interactions, when users receive unsatisfactory or inconsistent answers, they often provide an explicit corrective instruction, such as asking the model to ``look carefully,'' ``check the image again,'' or reconsider its answer based on the visual content. Turn-6 mimics this natural form of user intervention. It is not intended to measure autonomous self-recovery, since the model is explicitly guided to refocus on the image. Instead, it tests whether visual evidence remains latent and recoverable, or whether it has been irreversibly overwritten by the polluted textual history.

The observed second V-shaped pattern from Turn-5 to Turn-6 in Fig.~\ref{fig-V-shape} provides important evidence for this design. Across nearly all evaluated models, accuracy substantially increases at Turn-6 after severe degradation in previous turns. This universal recovery suggests that the visual information is not completely forgotten. Rather, it is progressively suppressed by accumulated hallucinated context and can be reactivated when the model is explicitly encouraged to re-anchor its reasoning to the image. In this sense, Turn-6 separates two different failure modes: loss of visual perception versus suppression of visual grounding by narrative inertia. The results support the latter explanation and provide direct motivation for our Conflict-Aware Visual Rectification method, which aims to restore visual grounding by refreshing visual representations and arbitrating conflicts between visual evidence and corrupted textual context.

\section{Details of Evaluation}
\label{app-evaluation}

MM-Snowball evaluates open-ended responses in multi-turn multimodal dialogues, where simple lexical matching or n-gram-based metrics are insufficient to capture semantic correctness. In particular, different turns impose different evaluation requirements: early turns mainly assess visual grounding, the adversarial turns require rejecting false or misleading premises, and the repair turn further examines whether the model can re-anchor its response to the visual evidence. Therefore, a reliable evaluator must judge not only semantic equivalence to the ground truth, but also whether the response satisfies the turn-specific objective.

To ensure consistent and scalable evaluation, we employ GPT-5 as a unified LLM judge. Given the image, dialogue context, current question, ground-truth answer, and model response, GPT-5 is instructed to determine whether the response is semantically consistent with the ground truth under the specific conversational setting. The evaluator is required to focus on factual correctness rather than superficial lexical overlap, and to explicitly penalize responses that follow hallucinated premises or contradict the image. The following is the detailed scoring protocol used for GPT-5 evaluation:
\begin{tcolorbox}[
  colback=lightgray!120,
  colframe=black,
  arc=0mm,
  boxrule=0.5pt,
  left=5pt,
  right=5pt,
  top=5pt,
  bottom=5pt,
  width=\textwidth,
]
\textbf{Prompt:}\\
Compare the following two texts: a reference text and a generated text. Determine whether the generated text correctly conveys the same meaning as the reference text. Do not require the wording to be identical—focus on whether the semantics, key facts, and intent are preserved. Minor differences in phrasing or style are acceptable. Respond with a judgment of correct if the generated text is semantically equivalent to the reference text; otherwise, it's wrong.

Output `yes' if correct, `no' if wrong. Output only `yes' or `no' without any additional text.
\end{tcolorbox}

We further validate the reliability of this automated evaluation protocol through human agreement analysis. Specifically, we compare GPT-5 judgments with annotations from three human experts and compute the Pearson correlation coefficient between LLM-based and human evaluation results. The correlation remains consistently high, with $r > 0.85$, indicating strong agreement between GPT-5 and human evaluators. This result confirms that the LLM-based evaluation provides a reliable and reproducible approximation of expert human judgment for MM-Snowball.

\section{Details of CAVR Method}
\label{app-cavr}

\subsection{Representation-level Visual Rectification (RVR)}

In deep multimodal Transformer architectures, the propagation of visual information often suffers from a phenomenon akin to \textit{visual fading}. As the autoregressive decoding penetrates deeper layers, the model's latent representations become increasingly dominated by linguistic priors and long-context dialogue histories, causing a significant attenuation of the original visual evidence. This visual fading disconnects the inference trajectory from the image grounding, rendering the model susceptible to hallucination snowballing.

To counteract this, we propose Representation-level Visual Rectification (RVR), a non-parametric, inference-time intervention mechanism. RVR conceptually reformulates the Feed-Forward Network (FFN) as a memory retrieval system. It dynamically monitors decoding uncertainty and, upon detecting ambiguity, executes a manifold-level visual retracing to realign the hidden states with the visual input.

\subsubsection{Entropy-based Epistemic Uncertainty Monitoring}
We posit that high epistemic uncertainty during token generation serves as a reliable proxy for visual-linguistic misalignment. To detect the onset of such ambiguity, we monitor the normalized entropy of the predictive distribution within the decoding trajectory. Let $\mathbf{h}_\ell \in \mathbb{R}^d$ denote the hidden state at layer $\ell$. The probability distribution over the vocabulary $\mathcal{V}$ is given by $\mathbf{p}_\ell = \text{Softmax}(\text{LMHead}(\mathbf{h}_\ell))$. We quantify the model's confusion using the normalized Shannon entropy:
\begin{equation}
\mathcal{U}_\ell = \frac{-\sum_{w \in \mathcal{V}} p_\ell(w) \log p_\ell(w)}{\log |\mathcal{V}|}
\end{equation}
By normalizing the entropy by the maximum possible entropy ($\log |\mathcal{V}|$), we map the uncertainty score to a scale-invariant range of $[0, 1]$. During autoregressive generation, if $\mathcal{U}_\ell$ exceeds a calibrated threshold $\tau$, it indicates that the current linguistic context is insufficient to resolve the next token deterministically, signaling a critical need to re-anchor the inference process with visual evidence.

\subsubsection{Manifold-level Visual Retracing}
Standard FFNs in Transformers can be theoretically interpreted as Key-Value memory networks, where the input hidden state retrieves information stored in the static weights. RVR extends this paradigm by introducing an auxiliary visual retracing pathway, allowing the hidden state to explicitly query the raw visual tokens when the static memory is insufficient.

Upon triggering the uncertainty alarm ($\mathcal{U}_\ell > \tau$), we intervene in the FFN of the layer. Instead of relying solely on the parametric linguistic memory, we execute a retrieval operation where the current hidden state $\mathbf{h}$ serves as a query to attend to the visual tokens. We define the visual rectification term $\Delta(\mathbf{h}, \mathbf{V})$ as a weighted aggregation of visual evidence based on semantic relevance:
\begin{equation}
\Delta(\mathbf{h}, \mathbf{V}) = \sum_{j=1}^{M} \phi(\langle \mathbf{h}, \mathbf{v}_j \rangle) \cdot \mathbf{v}_j
\end{equation}
where $\langle \cdot, \cdot \rangle$ denotes the inner product similarity, and $\phi(\cdot)$ is a non-linear activation function that normalizes the attention scores. This process effectively identifies and retrieves the visual features most relevant to the current semantic context.

The rectified output of the FFN, denoted as $\widetilde{\text{FFN}}$, is formulated as a linear interpolation between the standard linguistic processing and the visual retracing signal:
\begin{equation}
\widetilde{\text{FFN}}(\mathbf{h}) = (1 - \lambda) \cdot \text{FFN}(\mathbf{h}) + \lambda \cdot \Delta(\mathbf{h}, \mathbf{V})
\end{equation}
Here, $\lambda \in [0, 1]$ is the injection coefficient governing the intensity of visual reinforcement. By re-injecting raw visual features directly into the latent manifold, RVR effectively refreshes the model's short-term memory, correcting representation drift and ensuring that subsequent token generation remains grounded in the visual stimuli. Crucially, this operation leverages a lightweight retrieval mechanism compatible with pre-trained weights without requiring gradient updates.

\subsection{Logit-level Conflict Rectification (LCR)}
Standard autoregressive decoding in MLLMs is often plagued by contextual inertia, where the model prioritizes consistency with the dialogue history over fidelity to the visual input. This vulnerability is the primary driver of hallucination snowballing: once a factual error is introduced, the corrupted historical context acts as a source of noise that biases subsequent predictions. To sever this propagation chain, we propose Logit-level Conflict Rectification (LCR), a decoding-time intervention strategy that dynamically arbitrates between conversational continuity and visual truthfulness based on the principles of Residual Visual Decoding.

\subsubsection{Decoupling Contextual and Visual Streams}
To rigorously disentangle visual perception from contextual interference, we conceptualize the robust decoding process as the interplay of three distinct predictive streams computed in parallel at each timestep $t$.

\begin{itemize} [leftmargin=*, noitemsep, topsep=2pt]
    \item \textbf{Full Context Stream ($\mathbf{p}_{\text{full}}$):} Conditioned on the comprehensive context $\mathcal{C}_{\text{full}} = \{\mathcal{I}, \mathcal{H}, q_t\}$, this distribution maintains conversational flow but is susceptible to the inertia of historical errors.
    \item \textbf{Residual Visual Stream ($\mathbf{p}_{\text{res}}$):} Also termed the visual anchor, this distribution is conditioned solely on the image and current query $\mathcal{C}_{\text{res}} = \{\mathcal{I}, q_t\}$. By stripping away the dialogue history $\mathcal{H}$, it isolates the pristine visual evidence, serving as a factual reference unpolluted by previous hallucinations.
    \item \textbf{Linguistic Prior Stream ($\mathbf{p}_{\text{prior}}$):} Estimated from the query alone ($\mathcal{C}_{\text{prior}} = \{q_t\}$), this serves as a baseline to quantify the model's ``blind'' linguistic biases independent of visual stimuli.
\end{itemize}

By contrasting these distributions, we can mathematically isolate the true visual signal from both historical noise and pure language priors.

\subsubsection{Quantifying Perceptual Determinism}
To effectively arbitrate between the potentially corrupted historical context and the factual visual evidence, we must quantify the necessity of visual reliance for the current token. We posit that valid visual perception should manifest as a significant deviation from the model's inherent linguistic priors.

Following the adaptive distribution blending strategy in Residual Visual Decoding, we employ the Jensen-Shannon Divergence (JSD) to measure the conflict between the Residual Visual Stream ($\mathbf{p}_{\text{res}}$) and the Linguistic Prior Stream ($\mathbf{p}_{\text{prior}}$). We define this divergence as {Perceptual Determinism} $\tau$:
\begin{equation}
    \tau = \text{JSD}(\mathbf{p}_{\text{res}} \parallel \mathbf{p}_{\text{prior}}) = \frac{1}{2} \text{KL}(\mathbf{p}_{\text{res}} \parallel \mathbf{M}) + \frac{1}{2} \text{KL}(\mathbf{p}_{\text{prior}} \parallel \mathbf{M})
\end{equation}
where $\mathbf{M} = \frac{1}{2}(\mathbf{p}_{\text{res}} + \mathbf{p}_{\text{prior}})$ is the average distribution. Because JSD is symmetric and bounded in $[0, 1]$, it provides a robust, normalized metric for quantifying how much the visual input alters the model's prediction relative to a blind guess. A high $\tau$ indicates that the visual evidence provides a strong, decisive signal that contradicts the language prior, necessitating a shift towards the visual anchor.

\subsubsection{Conflict-Aware Logit Rectification}
Based on the quantified Perceptual Determinism $\tau$, LCR acts as a dynamic regulator to modulate the reliance on potentially corrupted dialogue histories. We introduce a \textit{rectification coefficient} $\lambda$ that determines the extent to which the model should shift its predictive distribution from the history-dependent stream towards the pristine residual visual stream. This coefficient is dynamically calibrated by a sensitivity hyperparameter $\beta$:
\begin{equation}
    \lambda = \min(\beta \cdot \tau, 1)
\end{equation}
The final decoding logits are synthesized through a conflict-aware linear interpolation between the full context logits and the visual residual logits:
\begin{equation}
    \mathbf{L}_{\text{final}} = (1 - \lambda) \cdot \mathbf{L}_{\text{full}} + \lambda \cdot \mathbf{L}_{\text{res}}
\end{equation}
where $\mathbf{L}$ denotes the output logits before Softmax. This rectification mechanism effectively functions as a ``hallucination circuit breaker'': when visual evidence is decisive (high $\tau$), $\lambda$ approaches 1, allowing the model to bypass the historical influence and prioritize the visual anchor, thereby correcting potential errors induced by the snowballing effect. Conversely, when the visual signal offers no distinct information ($\tau \approx 0$), the model gracefully falls back to the full context to maintain conversational fluency.

\section{Ablation Study on CAVR}
\label{app-ablation}
\begin{figure}[t]
    \centering
    \includegraphics[width=0.5\textwidth]{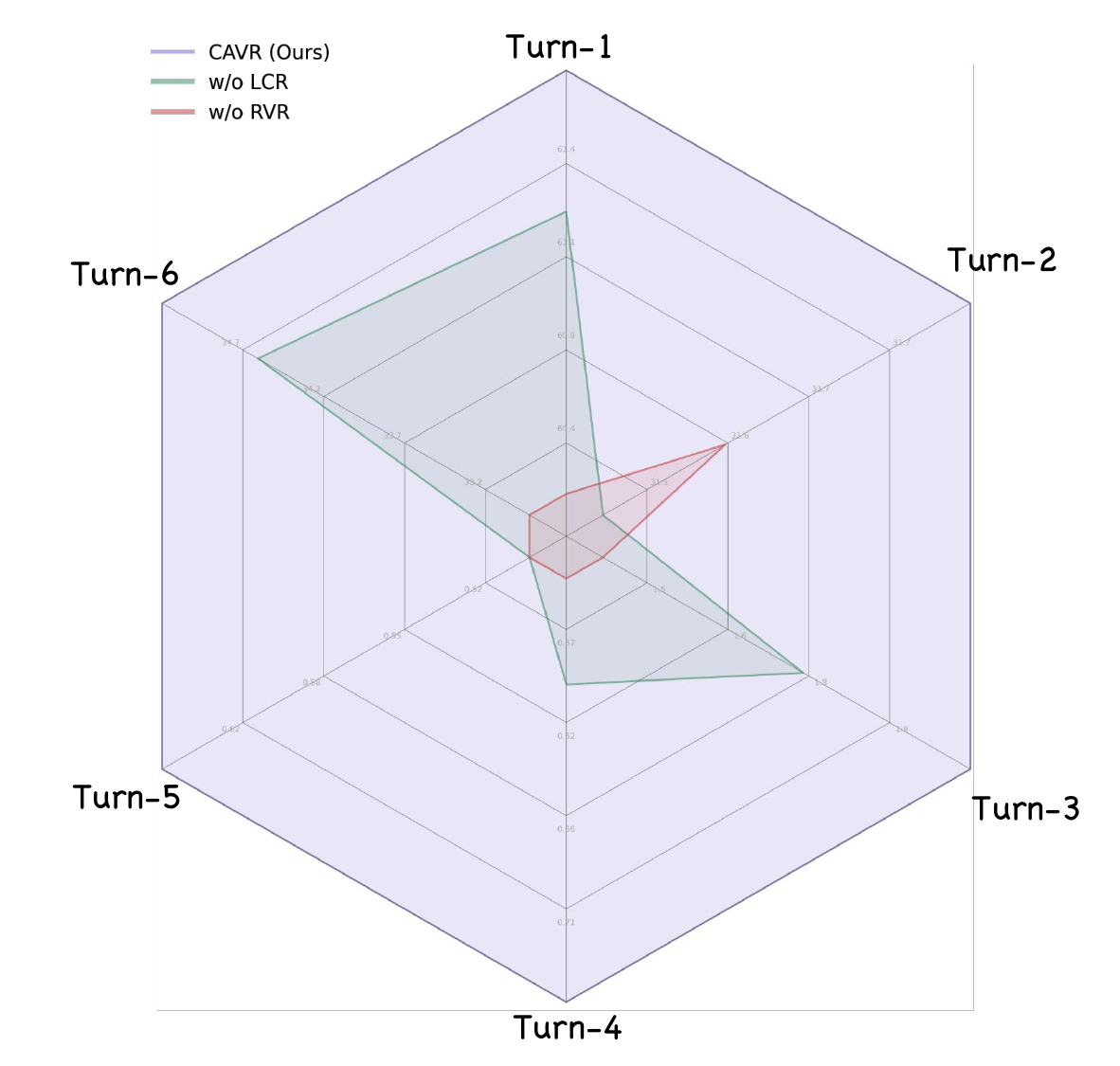}
    \caption{
\textbf{Ablation study for CAVR.}
The radar chart compares full CAVR with two ablated variants, w/o LCR and w/o RVR. CAVR consistently achieves the largest coverage across turns, demonstrating the complementary effect of Representation-level Visual Rectification and Logit-level Conflict Rectification.
}
    \label{fig-ab}
    \vskip -0.2in
\end{figure}

To evaluate the individual contribution of each component within the CAVR framework, we conduct an ablation study by isolating Representation-level Visual Rectification (RVR) and Logit-level Conflict Rectification (LCR). The performance across six dialogue turns is visualized in the radar chart in \cref{fig-ab}. We observe that the full CAVR model (represented by the outermost purple area) consistently outperforms the other two variants, indicating a strong synergistic effect between the internal representation alignment and external logit arbitration.

The most significant performance degradation occurs when RVR is removed (the ``w/o RVR'' red line). Without RVR, the model's internal hidden states rapidly lose their visual grounding as the dialogue depth increases, causing the performance to collapse, particularly from Turn 3 to Turn 6. This confirms our hypothesis that representation drift is a primary driver of the snowballing effect, and internal manifold-level retracing is indispensable for maintaining long-term visual fidelity. 

Similarly, the ``w/o LCR'' variant (the green line) also exhibits a consistent drop compared to the full model across all turns. Although RVR helps maintain better internal states, without LCR's logit-level filtering, the model remains vulnerable to linguistic inertia and corrupted historical context at the decision boundary. The fact that the purple polygon completely encloses the green and red areas demonstrates that RVR and LCR function complementarily: RVR ensures that the latent features remain visually informed, while LCR provides a robust circuit breaker to prevent history-induced noise from being decoded into final tokens.






\end{document}